\documentclass[conference]{IEEEtran}

\usepackage{graphicx}
\graphicspath{ {images/} }
\usepackage{cite}

\ifCLASSINFOpdf
\else
\fi

\usepackage[cmex10]{amsmath}
\usepackage{amsfonts}
\usepackage{bm}
\usepackage{xfrac}
\usepackage{mathtools}
\usepackage{array}
\usepackage{colortbl}
\usepackage{url}
\usepackage[hidelinks]{hyperref}

\usepackage{graphicx}  
\usepackage[caption=false,font=normalsize,labelfon
t=sf,textfont=sf]{subfig}

\usepackage{longtable}
\usepackage{booktabs}

\DeclareMathOperator*{\argmax}{arg\,max}

\begin{document}
%
\title{Drive-Net: Convolutional Network for Driver Distraction Detection}


\author{\IEEEauthorblockN{Mohammed S. Majdi$^{\dagger}$, Sundaresh Ram$^{\ast}$, Jonathan T. Gill$^{\dagger}$, Jeffrey J. Rodr\'iguez$^{\dagger}$}
	\IEEEauthorblockA{$^{\dagger}$Department of Electrical and Computer Engineering, University of Arizona, Tucson, AZ, USA\\$^{\ast}$School of Electrical and Computer Engineering, Cornell University, Ithaca, NY, USA\\ 		
		Email: \{{\tt \href{mailto:mohammadsmajdi@email.arizona.edu}{mohammadsmajdi},~\href{mailto:jtgill@email.arizona.edu}{jtgill},~\href{mailto:jjrodrig@email.arizona.edu}{jjrodrig}}\}{\tt \href{mailto:mohammadsmajdi@email.arizona.edu}{@email.arizona.edu}}, {\tt \href{mailto:sr2255@cornell.edu}{sr2255@cornell.edu}}}	
}

\maketitle

\begin{abstract}
 To help prevent motor vehicle accidents, there has been significant interest in finding an automated method to recognize signs of driver distraction, such as talking to passengers, fixing hair and makeup, eating and drinking, and using a mobile phone. In this paper, we present an automated supervised learning method called Drive-Net for driver distraction detection. Drive-Net uses a combination of a convolutional neural network (CNN) and a random decision forest for classifying images of a driver. We compare the performance of our proposed Drive-Net to two other popular machine-learning approaches: a recurrent neural network (RNN), and a multi-layer perceptron (MLP). We test the methods on a publicly available database of images acquired under a controlled environment containing about 22425 images manually annotated by an expert. Results show that Drive-Net achieves a detection accuracy of 95\%, which is 2\% more than the best results obtained on the same database using other methods.
\end{abstract}


\IEEEpeerreviewmaketitle

\section{Introduction}
\label{sec:intro}

\begin{figure}[!t]
	\centering
	\includegraphics[width=\textwidth/2]{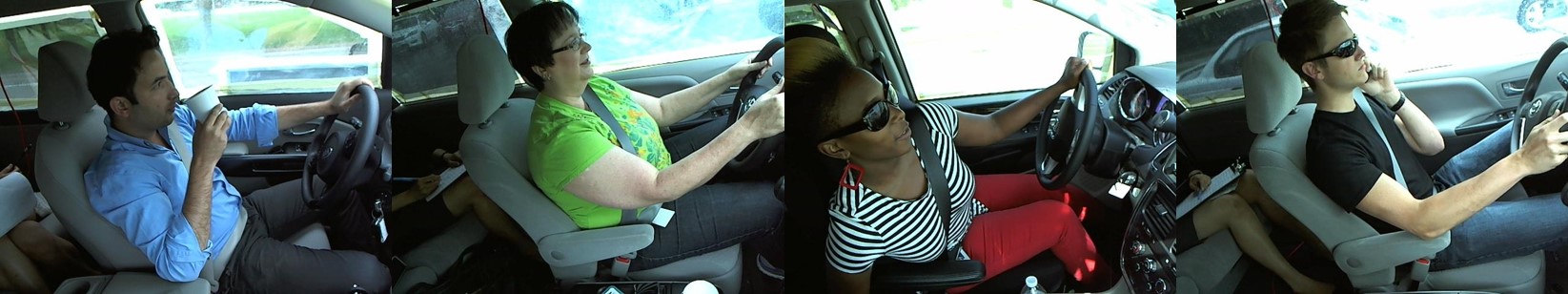}
	\caption{Some representative dashboard camera images of drivers being distracted. \emph{From left to right:} drinking while driving, safe driving, reaching behind when driving, talking on the phone (left hand) while driving \cite{StateFarm}.}
\label{fig1}
\vspace{-3mm}
\end{figure}

Distracted driving is a major cause of motor vehicle accidents. Each day in the United States, approximately 9 people are killed and more than 1000 are injured in crashes that involve a distracted driver \cite{NCSADD}. It is estimated that roughly 25\% of motor vehicle accident fatalities are due to distracted driving \cite{InsuranceJournal}. A study of American motor vehicle fatalities \cite{NCSADD} reveals the top 10 causes of distracted driving:
\begin{enumerate}
	\item	Generally distracted or ``lost in thought” -– 62\%
	\item Mobile phone use –- 12\%
	\item Outside person, object, or event -- 7\%. 
	\item Other occupants -– 5\%. 
	\item Using or reaching for a device brought into the car (e.g., phone) – 2\%. 
	\item Eating or drinking -– 2\%. 
	\item Adjusting audio or climate controls –- 2\%. 
	\item Using devices/controls to operate the vehicle (e.g., adjusting mirrors or seatbelts) –- 1\%.
	\item Moving objects (e.g., insects or pets) -– 1\%
	\item Smoking related –- 1\%.
\end{enumerate}
Therefore, there is an interest in using analysis of dashboard camera images to automatically detect drivers engaged in distracting behavior. A dataset of such dashboard camera images, observing various activities of drivers, has been compiled and used for the Kaggle competition regarding automated detection of driver distraction \cite{StateFarm}. Fig.~\ref{fig1} shows examples of a few dashboard camera images manually annotated as different activities when driving.

Object detection and human behavior detection are well-researched topics in the computer vision literature \cite{borji14}. Machine learning (esp. deep learning) techniques can often learn complex models and achieve high accuracy, so many researchers have started to apply such techniques to solve computer vision problems including object detection and human behavior detection. For example, the Inception-v4 model proposed by Szegedy \emph{et al.} \cite{szegedy} is a supervised learning model made up of deep convolutional residual networks (ResNet) that has more than 75 trainable layers, and it has been shown to achieve 96.92\% accuracy on the ImageNet dataset. Girshick \emph{et al.} \cite{Girshick} introduced a very powerful method for object detection and segmentation using a region-based convolutional neural network (CNN). This method divides the human behavior detection problem into two problems. First, they apply an object detection algorithm to detect the regions of interest (ROI) where people are present within an image. Next, each ROI is fed to a CNN to identify the type of behavior exhibited in the given ROI. Adding other traditional machine learning methods such as ensemble learning (i.e., bagging), and K nearest neighbors (KNN) to the CNN model is a way of improving the accuracy of the already existing model \cite{kim17}. 
\begin{figure*}[!t]
	\centering
	\includegraphics[width = \textwidth/1]{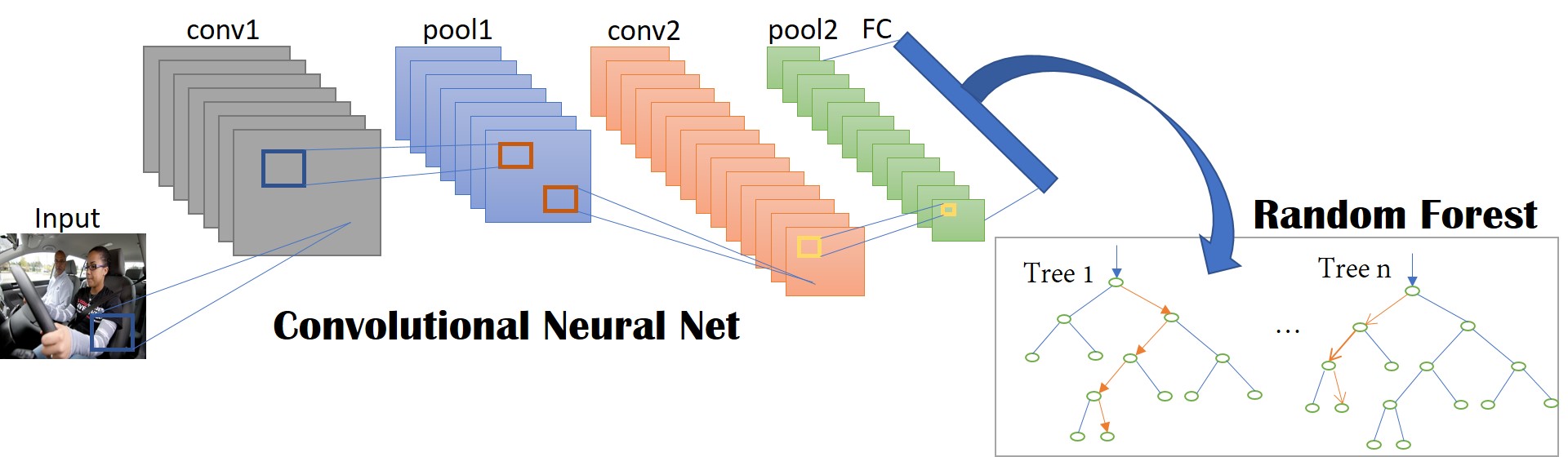}
	\caption{An overview of the proposed Drive-Net. Our proposed CNN architecture (shown on the left side) consists of two convolution layers (conv) each followed by a maxpooling layer (pool) and a final ReLU layer, the output of which is regularized using dropouts to obtain a fully connected layer (FC). The FC layer is fed as input to the random forest classifier (on the right side), which predicts the final class label.}
	\label{CNN}
	\vspace{-2mm}
\end{figure*}

One of the main drawbacks of CNNs is that training the network using a large dataset can lead to over-fitting the model. In order to avoid this, ensemble methods such as random decision forests can be effective. With this in mind, we propose a new supervised learning algorithm called Drive-Net that combines a CNN and a random forest in a cascaded fashion for application to the problem of driver distraction detection using dashboard camera images. We compare our proposed Drive-Net to two other neural network methods: a residual neural network (RNN) and a multi-layer perceptron (MLP). We show that Drive-Net achieves better classification accuracy than the driver distraction detection algorithms that were proposed in the Kaggle competition\cite{StateFarm}.

\section{Methods}

Our proposed method, Drive-Net, is a cascaded classifier consisting of two stages: a CNN as the first stage, whose output layer is fed as the input to a random decision forest to predict the final class label. We define each stage in detail below.

\subsection{Convolutional Neural Network Configuration} 

We adopt the U-Net architecture \cite{ronneberger15} as the basis of our CNN. The motivation behind this architecture is that the contracting path captures the context around the objects in order to provide a better representation of the object as compared to architectures such as Alexnet \cite{krizhevsky12} and VGGnet \cite{Simonyan}. Very large networks like Alexnet and VGGnet require learning a massive number of parameters and are very hard to train in general, needing significant computational time. Thus, we empirically modify the U-Net architecture in this work to suit our application.

To construct our CNN, we discard U-Net's layers of up-convolution and the last two layers of down-sampling and replace them with a $1 \times 1$ convolution instead to obtain a fully connected layer. We use the rectifier activation function \cite{ronneberger15} for our CNN as the constant gradient of rectified linear units (ReLUs) results in faster learning and reduces the problem of vanishing gradient compared to hyperbolic tangent (tanh). We implement a maxpooling layer instead of average pooling in the sub-sampling layer \cite{krizhevsky12}. We observed that the performance is better when a ReLU layer was configured with maxpooling layer, resulting in higher classification accuracy after 50 epochs. We used the $1 \times 1$ convolutional filters for the Adam \cite{kingma15} optimizer. All the other parameters such as number of layers, convolutional kernel size, training algorithm, and the number of neurons in the final dense layer were all experimentally determined for our application. 

In order to keep the training time small, we reduce the size of dashboard camera images by a factor of 10 making them $64 \times 48$ in size and feed these as the input to our CNN. We do not zero-pad the image patches, as the ROI with human activity is located towards the center of the image. Two consecutive convolutional layers are used in the network. The first convolutional layer consists of 32 kernels of size $5 \times 5 \times 1$. The second convolutional layer consists of 64 kernels of size $5 \times 5 \times 32$. The sub-sampling layer is set as the maximum values in non-overlapping windows of size $2 \times 2$ (stride of 2). This reduces the size of the output of each convolutional layer by half. After the two convolutional and sub-sampling layers, we use a ReLU layer, where the activation $y$ for a given input $x$ is obtained as
\begin{eqnarray}
y = f(x) = \text{max}(0,x)
\label{eqn1}
\end{eqnarray}
A graphical representation of the architecture of the proposed CNN model is shown in Fig.\,\ref{CNN} (see the left side).

\subsection{Random Decision Forest}

A random forest classifier consists of a collection of decision tree classifiers combined together to predict the class label, where each tree is grown in some randomized fashion. Each decision tree classifier consists of decision (or split) nodes and prediction (or leaf) nodes. The prediction nodes of each tree in the random forest classifier are labeled by the posterior distribution over the image classes \cite{bosch07}. Each decision node contains a test that splits best the space of data to be classified. An image is classified by sending it down the decision tree and aggregating the reached leaf posterior distributions. Randomness is usually injected at two points during training: when sub-sampling the training data and when selecting node tests. Each tree within the random forest classifier is binary and grown in a top-down manner. We choose the binary test at each node by maximizing the information gain,
\begin{eqnarray}
\Delta E = -\sum_{i}\frac{\mid Q_{i}\mid}{\mid Q\mid}E(Q_{i})
\label{eqn2}
\end{eqnarray} 
obtained by partitioning the training set $Q$ of image patches into two sets $Q_{i}$ according to a given test. Here $E(q)$ is the entropy of the set $q$ and $\mid \cdot\mid$ is the size of the set. We repeat this selection process for each decision node until it reaches a certain depth. Many implementations of random forests \cite{winn06, lepetit06} use simple pixel-level tests at the nodes because it results in faster tree convergence. As we are interested in features that encode the shape, and appearance, we are interested in spatial correspondence between the pixels. Therefore, we use a simple test proposed by \cite{bosch07} --- a linear classifier on the feature vector --- at each decision node.

Suppose $T$ is the set of all trees, $C$ is the set of all classes, and $L$ is the set of all leaves for a given tree. During training, the posterior probabilities $\left(P_{t,l}(Y(I) = c)\right)$ for each class $c \in C$ at each leaf node $l \in L$, are found for each tree $t \in T$. These probabilities are calculated as the ratio of the number of images $I$ of class $c$ that reach a leaf node $l$ to the total number of images that reach that leaf node $l$. $Y(I)$ is the class label for image $I$. During test time, we pass a new image through every decision tree until it reaches a prediction (or leaf) node, average all the posterior probabilities, and classify the image as
\begin{eqnarray}
\hat{Y}(I) = \argmax_{c}\left\{\frac{1}{\mid T\mid}\sum_{t=1}^{\mid T\mid} P_{t,l}(Y(I) = c)\right\}
\label{eqn4}
\end{eqnarray}
where $l$ is the leaf node reached by the image $I$ in tree $t$. A graphical representation of the proposed random forest classifier is shown in Fig.\,\ref{CNN} (see the right side).

\section{Experiments and Results}

\subsection{Dataset}

The Kaggle competition \cite{StateFarm} for driver distraction has provided 22425 images for training and 79727  for test. Since we did not have access to the test labels, our experiments were done solely on the training images. However, the quality and conditions of training and testing images are similar; the only difference is that none of the drivers used in the training dataset appear in images in the test dataset. The images are of size $640 \times 480$, and for our experiments we converted them from color to grayscale.

There are ten classes provided, related to the ones listed in Section~\ref{sec:intro}. Each class includes almost tens of the  data, so that we have a uniform distribution of sample data.
\begin{itemize}
	\item c0: safe driving
	\item c1: texting (right hand)
	\item c2: talking on the phone (right hand)
	\item c3: texting (left hand)
	\item c4: talking on the phone (left hand)
	\item c5: operating the radio
	\item c6: drinking
	\item c7: reaching behind
	\item c8: hair and makeup
	\item c9: talking to passenger
\end{itemize}

\subsection{Algorithm Parameters}

The convolutional neural random forest classifier is implemented using TensorFlow \cite{abadi15}, and runs on an NVIDIA GeForce GTX TITAN X GPU with 16GB of memory. The classifier was trained using the stochastic gradient descent algorithm, Adam \cite{kingma15}, to efficiently optimize the weights of the CNN. The weights were normalized using initialization as proposed in \cite{kingma15} and updated in a mini-batch scheme of 128 candidates. The biases were initialized with zero, and the learning rate was set to $\alpha = 0.001$. The exponential decay rates for the first- and second-moment estimates were set as $\beta_{1} = 0.9$ and $\beta_{2} = 0.99$, respectively. We used $\epsilon = 10^{-8}$ to prevent division by zero. A dropout rate of 0.5 was implemented as regularization, applied to the output of the last convolutional layer and the dense layer to avoid overfitting. Finally, we used an epoch size of 50. The softmax loss (cross-entropy error loss) was utilized to measure the error loss. We used 100 estimators and a keep rate of $\gamma = 10^{-4}$ for the random forest algorithm. 

\subsection{Performance Evaluation}

We tested the algorithm performance by conducting a \emph{k}-fold cross validation on the entire dataset. For our experiments we varied values of \emph{k}$\in [2, 5]$ and found that the results were consistent enough to indicate that the network is not over-fitting. Hence, we chose \emph{k} = 5. First, we randomized the order of the driver images within the dataset. For each fold of the \emph{k}-fold cross validation we chose 80\% of the total 22425 images as the training dataset and tested the trained model on the remaining 20\% of the images. We made sure that the images from the entire dataset appeared in the test dataset only once in all of the \emph{k}-folds, thereby allowing each image to be classified as a test image exactly once. 

We compared our proposed Drive-Net with two other neural network classifiers: a RNN classifier \cite{liang15}, and a MLP classifier\cite{haykin09}. We report the classification accuracy, which is defined as the percentage of correct predictions and the number of false positives (a.k.a. false detections) for each class as the figures of merit for comparing the algorithms. For classification accuracy, we present the results of seven other methods based on support vector machines (SVMs), dimensionality reduction techniques such as principal component analysis (PCA), feature extraction techniques such as histogram of oriented gradients (HOG), very deep convolutional nets such as VGG-16, VGG-GAP and an ensemble of these two as reported by Zhang in \cite{Yundong} using the same Kaggle dataset of 22425 images. 

Table~\ref{table1} shows the mean classification accuracy of the different classifiers as reported in Zhang \cite{Yundong} and that of the three neural network classifiers that we implemented. From Table~\ref{table1}, we observe that the Drive-Net achieves a classification accuracy of 4.8 percentage points greater than the VGG-16 classifier, 3.7 percentage points greater than a VGG-GAP classifier, 2.4 percentage points greater than an ensemble of VGG-16 and VGG-GAP classifier, 3.3 percentage points greater than the RNN classifier and 13 percentage points greater than the MLP classifier. 

Table~\ref{table2} shows the number of false classifications for our Drive-Net and for the RNN and MLP. From Table~\ref{table2}, we observe that our Drive-Net is able to identify the classes c6 (drinking) and c3 (texting with left hand) with minimum false detections, whereas the RNN and MLP classifiers have a hard time distinguishing these classes with many false detections, usually higher than the number of false detections in the other classes of these methods. Also, the total number of false detections for our Drive-Net is an order of magnitude smaller than that of the MLP classifier and slightly smaller in comparison to the RNN classifier.  

\begin{table}[!t]
\caption{Mean classification accuracy of the automated methods}
\label{table1}
  \begin{center}
	\renewcommand{\arraystretch}{1.4}
	\begin{tabular}{>{\centering} m{5cm} >{\centering} m{3.3cm}} 
	\hline
	\rowcolor[gray] {0.8}\textbf{Method} & \textbf{Accuracy}\tabularnewline \hline
	\textbf{Methods from \cite{Yundong}:} \tabularnewline
	Pixel SVC & 18.3\%  \tabularnewline 
	SVC + HOG & 28.2\% \tabularnewline
	SVC + PCA & 34.8\% \tabularnewline
	SVC + Bbox + PCA & 40.7\% \tabularnewline
	VGG-16 & 90.2\% \tabularnewline
	VGG-GAP & 91.3\% \tabularnewline
	Ensemble VGG-16 and VGG-GAP & 92.6\% \tabularnewline \hline
	\textbf{Methods we implemented:} & \tabularnewline
	MLP & 82\textcolor{white}{.0}\% \tabularnewline
	RNN & 91.7\% \tabularnewline
	Drive-Net & 95\textcolor{white}{.0}\% \tabularnewline
	\hline
	\end{tabular}
   \end{center}
\vspace{-4mm}
\end{table}

\section{Conclusion}

Distracted driving is a major cause of motor vehicle accidents. Therefore, there is a significant interest in finding automated methods to recognize signs of driver distraction from dashboard camera images installed in vehicles. We propose a solution to this problem using a supervised learning framework. Our method named Drive-Net combines a CNN and a random forest classifier to recognize the various driver distraction categories in images. We apply our Drive-Net to a publicly available dataset of images used in a Kaggle competition and show that our Drive-Net achieves better accuracy than the driver distraction algorithms reported in the competition. We also compared Drive-Net to two other neural network algorithms: a RNN and a MLP algorithm, using the same dataset. The results show that Drive-Net achieves better detection accuracy compared to the other two algorithms.

\begin{table}[!t]
\caption{The error count of each class for the neural network methods}
\label{table2}
  \begin{center}
	\renewcommand{\arraystretch}{1.4}
	\begin{tabular}{>{\centering} m{1.65cm} >{\centering} m{1.65cm}  >{\centering}m{1.65cm}  >{\centering}m{1.65cm}} 
	\hline
	\rowcolor[gray] {0.8}\textbf{Class} & \textbf{Drive-Net} & \textbf{MLP} & \textbf{RNN}\tabularnewline \hline
	c0 & 35 & 356 & 48 \tabularnewline 
	c1 & 17 & 199 & 34 \tabularnewline
	c2 & 14 & 158 & 31 \tabularnewline
	c3 & 09 & 116 & 47 \tabularnewline
	c4 & 34 & 252 & 30 \tabularnewline
	c5 & 15 & 108 & 14 \tabularnewline
	c6 & 08 & 263 & 18 \tabularnewline
	c7 & 21 & 117 & 10 \tabularnewline
	c8 & 29 & 181 & 46 \tabularnewline
	c9 & 26 & 268 & 46 \tabularnewline \hline
	All Classes & 208 & 2018 & 324 \tabularnewline
	\hline
	\end{tabular}
   \end{center}
\vspace{-4mm}
\end{table}
\end{document}